\documentclass{article}

\usepackage{arxiv}

\usepackage[utf8]{inputenc} 
\usepackage[T1]{fontenc}    
\usepackage[pagebackref,breaklinks,colorlinks]{hyperref}
\usepackage{url}            
\usepackage{booktabs}       
\usepackage{amsfonts}       
\usepackage{nicefrac}       
\usepackage{microtype}      
\usepackage{lipsum}
\usepackage{graphicx}

\usepackage{orcidlink}

\usepackage{cleveref}
\usepackage{comment}
\usepackage{gensymb}
\usepackage{pifont}

\usepackage{siunitx}

\usepackage{xspace}
\makeatletter
\DeclareRobustCommand\onedot{\futurelet\@let@token\@onedot}

\def\@onedot{\ifx\@let@token.\else.\null\fi\xspace}

\def\eg{\emph{e.g}\onedot} 
\def\ie{\emph{i.e}\onedot}

\def\etal{\emph{et al}\onedot}
\makeatother

\graphicspath{ {./images/} }

\title{Domain Generalization for 6D Pose Estimation Through NeRF-based Image Synthesis}

\author{ \href{https://orcid.org/0000-0002-4041-0448}{\includegraphics[scale=0.06]{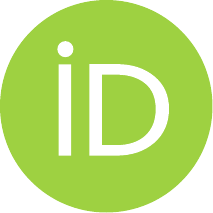}\hspace{1mm}Antoine~Legrand} \\
	Dept. of Electrical Engineering,\\
        ICTEAM, UCLouvain\\
	Dept. of Electrical Engineering, \\ 
        ESAT, KU Leuven\\
        Aerospacelab \\
	\texttt{antoine.legrand@uclouvain.be} \\
	\And
	\href{https://orcid.org/0000-0003-0597-1167}{\includegraphics[scale=0.06]{orcid.pdf}\hspace{1mm}Renaud~Detry} \\
	Dept. of Electrical Engineering, \\ 
        ESAT, KU Leuven\\
	Dept. of Mechanical Engineering, \\ 
        MECH, KU Leuven\\
	\AND
	\href{https://orcid.org/0000-0001-5049-2929}{\includegraphics[scale=0.06]{orcid.pdf}\hspace{1mm}Christophe~De~Vleeschouwer} \\
	Dept. of Electrical Engineering,\\
        ICTEAM, UCLouvain\\
}

\begin{document}
\maketitle
\begin{abstract}

    This work introduces a novel augmentation method that increases the diversity of a train set to improve the generalization abilities of a 6D pose estimation network. For this purpose, a Neural Radiance Field is trained from synthetic images and exploited to generate an augmented set. Our method enriches the initial set by enabling the synthesis of images with (i) unseen viewpoints, (ii) rich illumination conditions through appearance extrapolation, and (iii) randomized textures. 
    We validate our augmentation method on the challenging use-case of spacecraft pose estimation and show that it significantly improves the pose estimation generalization capabilities. On the SPEED+ dataset, our method reduces the error on the pose by $50\%$ on both target domains. 
    
\end{abstract}

\section{Introduction}
\label{sec_intro}

    Deep neural networks require large amount of data for their training. Due to the cost and complexity of acquiring and annotating real-world images, synthetic images have become a \textit{de facto} standard to train them. However, synthetic images do not perfectly capture the real world. For example, rendering tools struggle to accurately emulate the illumination conditions encountered in the wild. Similarly, the realism of objects rendered by these tools is highly dependent on the quality of their virtual 3D model. Those mismatches between the synthetic images and their real counterpart are responsible for the domain shift problem, \ie, the drop in accuracy encountered by a network trained on a source domain, \eg, synthetic images, when it is used for prediction on a target domain, \eg, real images.
    
    This domain shift problem can be mitigated by either domain adaptation~\cite{farahani2021brief}, \ie, techniques that aim at bringing the train set distribution closer to the real one, or domain generalization~\cite{wang2022generalizing}, \ie, techniques that aim at enlarging the training set distribution to enforce the network to learn domain-invariant features and therefore generalize to unseen domains. Because they do not assume the knowledge of the target domain, domain generalization techniques can be applied on a broader range of tasks. In this paper, we introduce a domain generalization technique for the task of estimating the relative 6D pose, \ie, position and orientation, of a given object.

\begin{figure}[t]
    \centering
    \includegraphics[width=0.99\linewidth]{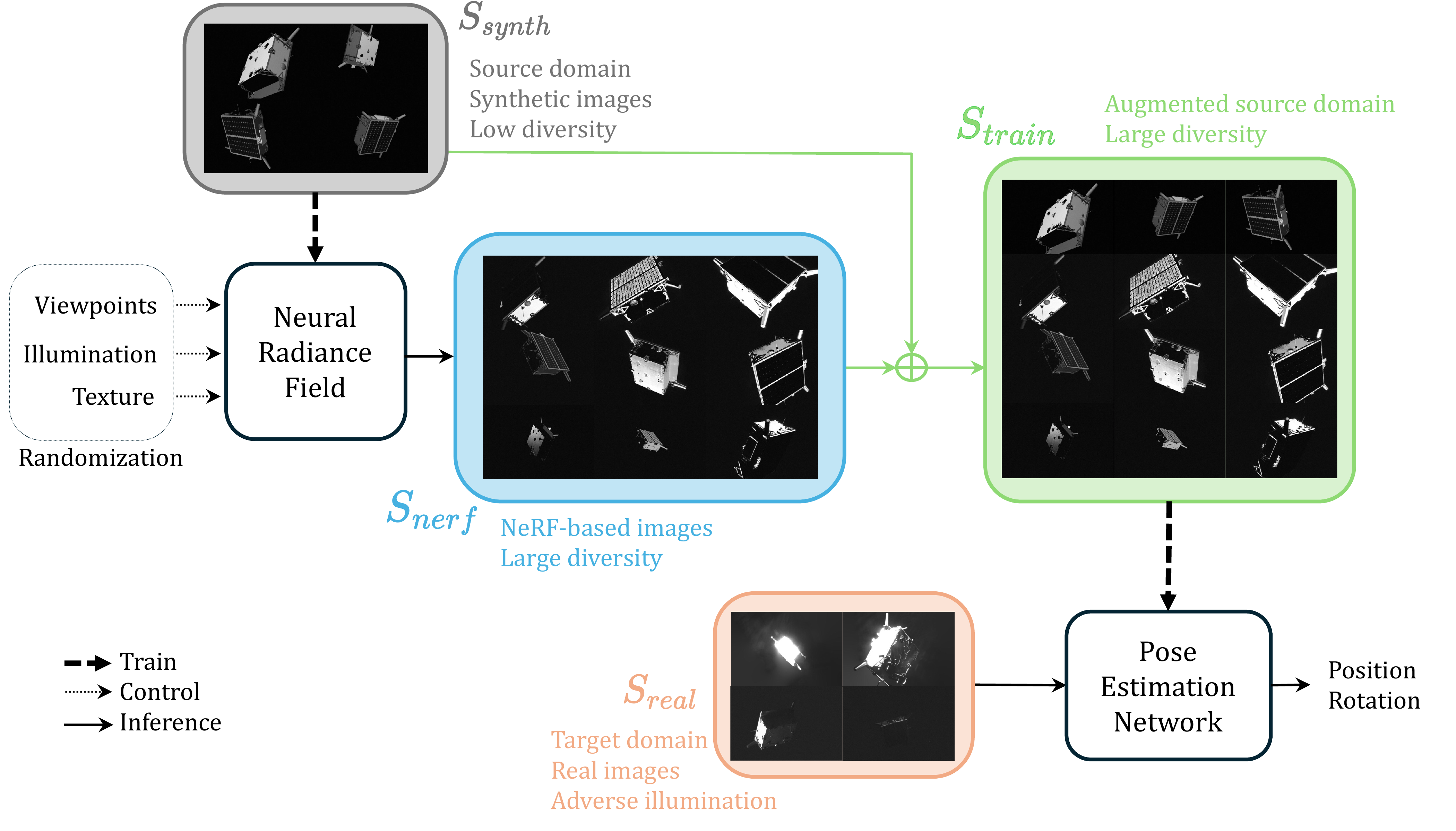}
    \caption{\label{fig_overview_intro} Overview of our domain generalization method. Instead of training a pose estimation network on a synthetic set $S_{synth}$, which lacks diversity, we train it on an augmented set $S_{train}$ that combines the synthetic set and $S_{nerf}$, a set of images synthesized by a Neural Radiance Field~\cite{mildenhall2021nerf} (NeRF) trained on synthetic images. Our proposal is shown to improve the accuracy of the pose estimation module on the real test set $S_{real}$. The diversity of the generated set, $S_{nerf}$, is ensured by randomizing the viewpoints distribution, the illumination conditions as well as the target texture, as further explained in \Cref{sec_nerf_augm}.}
\end{figure} 
    
    Our method relies on the augmentation of a synthetic set through a Neural Radiance Field (NeRF)~\cite{mildenhall2021nerf} to improve the generalization abilities of a pose estimation network. As depicted in \Cref{fig_overview_intro}, a NeRF is trained using a synthetic set $S_{synth}$. This enables the generation by the NeRF of a novel set $S_{nerf}$ that is diverse in terms of poses distribution, illumination conditions and textures. Finally, the pose estimation network is trained on a train set $S_{train} = S_{synth} \cup S_{nerf}$ that is more diverse than the original synthetic set $S_{synth}$ as it combines this synthetic set with the NeRF-based one, $S_{nerf}$. This increase in training set diversity results in enhanced domain generalization abilities for the pose estimator.

\begin{figure}[b]
    \centering
    \includegraphics[width=0.99\linewidth]{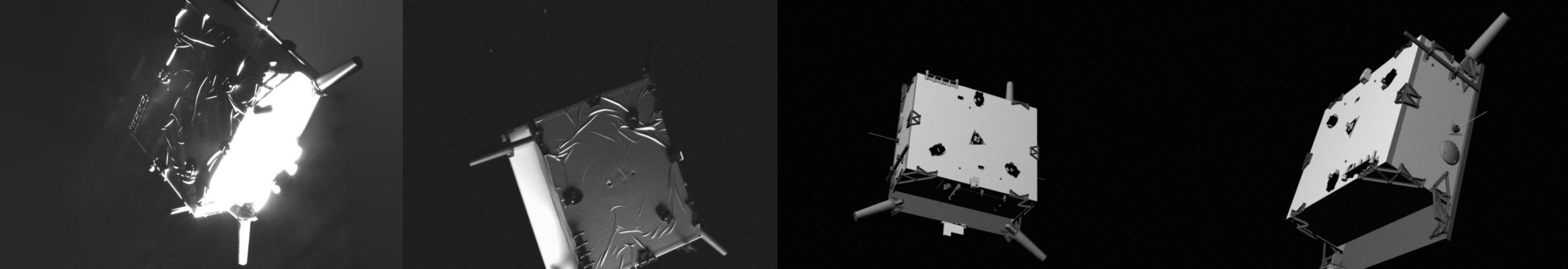}
    \caption{\label{fig_mismatches_domain} \textbf{(Left)} Two real images depicting the spacecraft used in the SPEED+ dataset~\cite{park2022speed+}. \textbf{(Right)} Two synthetic images from SPEED+, generated using a simplified CAD model of the target spacecraft. The global shape of the spacecraft is correctly rendered but the synthetic images fail to capture its texture. In addition, the synthetic images do not contain the adverse illumination conditions encountered on the real ones.}
\end{figure} 
    
    We demonstrate the interest of our method on the on-orbit 6D pose estimation task, \ie, arguably one of the most challenging domain generalization use-cases. This task aims at estimating the 6D pose of an uncooperative, known, target spacecraft, relative to a monocular camera, a key capability for future autonomous space missions such as on-orbit servicing~\cite{henshaw2014darpa}. This task is especially challenging from the domain generalization point-of-view. It is currently addressed through pose estimation networks solely trained on synthetic images 
    ~\cite{chen2019satellite,proencca2020deep,park2024robust} since the cost related to the acquisition of images depicting the target in orbit is prohibitive. Unfortunately, as illustrated in \Cref{fig_mismatches_domain}, the synthetic images poorly capture the illumination conditions encountered on-orbit. The absence of atmospheric diffusion is responsible for large contrasts in the image while the direct illumination of the Sun on the specular components of the target results in over-exposed spacecraft parts. Moreover, the CAD model used to render the target is simplified compared to the actual spacecraft. As a result, the spacecraft presented in the synthetic images does not reflect the real target appearance distribution.

    Our contribution therefore consists in a novel data augmentation method for improving the generalization abilities of a 6D pose estimation network trained from those augmented data. It relies on the augmentation of a synthetic train set through a Neural Radiance Field, exploited to synthesize novel training images that increase the diversity of the train set in terms of pose distribution, illumination conditions and texture. This enlarged training set results in enhanced generalization abilities for the pose estimation network. 
    We show that the method can be used to improve the generalization abilities of a pose estimator. Ablation studies assess the impact of the appearance and texture augmentations on the generalization abilities. 
    They also show that the set of images generated with the NeRF is sufficient to learn a robust pose estimator. This paves the way for training a pose estimator in scenarios where the NeRF would be trained differently than with a set of CAD-based synthetic images, \eg, based on a small set of real images as done in~\cite{legrand2024leveraging}.  
    The rest of the paper is organized as follows. \Cref{sec_related_works} positions our work with respect to previous works. \Cref{sec_method} describes our proposed method, while \Cref{sec_experiments} validates it. \Cref{sec_conclusion} concludes.

\section{Related Works}
\label{sec_related_works}

    The domain shift, \ie, the mismatch between source domain(s) and target domain(s), is encountered in many computer vision applications~\cite{kondrateva2021domain,sankaranarayanan2018learning} and therefore received significant interest in the previous years. Current works focus on either domain adaptation~\cite{farahani2021brief} or domain generalization~\cite{wang2022generalizing}. While the former exploit priors on the target domain(s) to train the network, the latter rely only on the source domain(s). Due to their broader range of applicability, we focus on domain generalization approaches. They either rely on multiple source domains or on one. Domain alignment techniques~\cite{muandet2013domain} aims at aligning the features predicted on these different source domains. In ensemble learning~\cite{sagi2018ensemble}, different models are learned on different domains, and the inference on the target domain is conducted using the ensemble of the trained models. Meta-learning frameworks~\cite{hospedales2021meta} rely on pseudo-train and pseudo-test sets taken from different source domains and optimize the network so that its accuracy on the pseudo-test sets increases. Although these techniques have demonstrated their interest, they rely on multiple source domains. Data augmentation techniques also address this domain generalization problem. For example, Crossgrad~\cite{shankar2018generalizing} or L2A-OT~\cite{zhou2020learning} generates novel training data that improves the generalization abilities of the network. MixStyle~\cite{zhou2024mixstyle} follows a similar goal but performs the augmentation at the feature level by mixing the features of different source domains. Regarding the single domain generalization problem~\cite{qiao2020learning}, data augmentation or generation techniques have been considered. They aim at synthesizing novel domains from the available source domain, thereby enforcing the network to learn domain-invariant features. Previous works~\cite{tobin2017domain,wang2021learning} highlighted that increasing the diversity of a training set through data augmentation improves the generalization abilities of a network trained on this set. 

    Our work focuses on the spacecraft pose estimation task. It aims at predicting the position and orientation of a target spacecraft relative to a monocular camera~\cite{kisantal2020satellite}. This task encounters a significant domain shift problem. Indeed, due to the lack of available real images depicting the target in-orbit, pose estimators are solely trained on synthetic images. Moreover, they do not capture the adverse illumination conditions encountered on orbit, nor the actual spacecraft texture, as illustrated in \Cref{fig_mismatches_domain}. Hence, without a proper learning strategy, pose estimators trained on such synthetic images perform poorly on real images~\cite{kisantal2020satellite,park2023spec2021}. Although domain adaptation techniques have been considered on this task~\cite{park2023spec2021,perez2022spacecraft,wang2023bridging}, domain generalization techniques are preferred because they require no prior on the target domain. Several works~\cite{park2024robust,cassinis2022ground,park2023spec2021,ulmer20236d,legrand2024domain} addressed the problem through data augmentation techniques such as Brightness \& Contrast augmentation, random erasing~\cite{zhong2020randomerasing}, Hide \& Seek~\cite{singh2018hide}, Postlamp~\cite{sakkos2019illumination}, neural style augmentation~\cite{jackson2019style} or Fourier-based augmentations~\cite{xu2023fourier}. They highlighted the fact that increasing the diversity of the train set results in stronger generalization abilities~\cite{legrand2024domain}. Our work builds on this conclusion to address the domain shift by increasing the diversity of a synthetic train set. Instead of relying on classical, \ie, image-level, augmentations, we augment the whole training set through a Neural Radiance Field~\cite{mildenhall2021nerf} (NeRF) trained on the original train set. By randomizing the image synthesis process of the NeRF, we massively increase the diversity of the train set in terms of viewpoints distribution, illumination, and texture conditions, which, in turn, enhances the generalization abilities of a pose estimator.

\begin{figure}[t]
    \centering
    \includegraphics[width=0.99\linewidth]{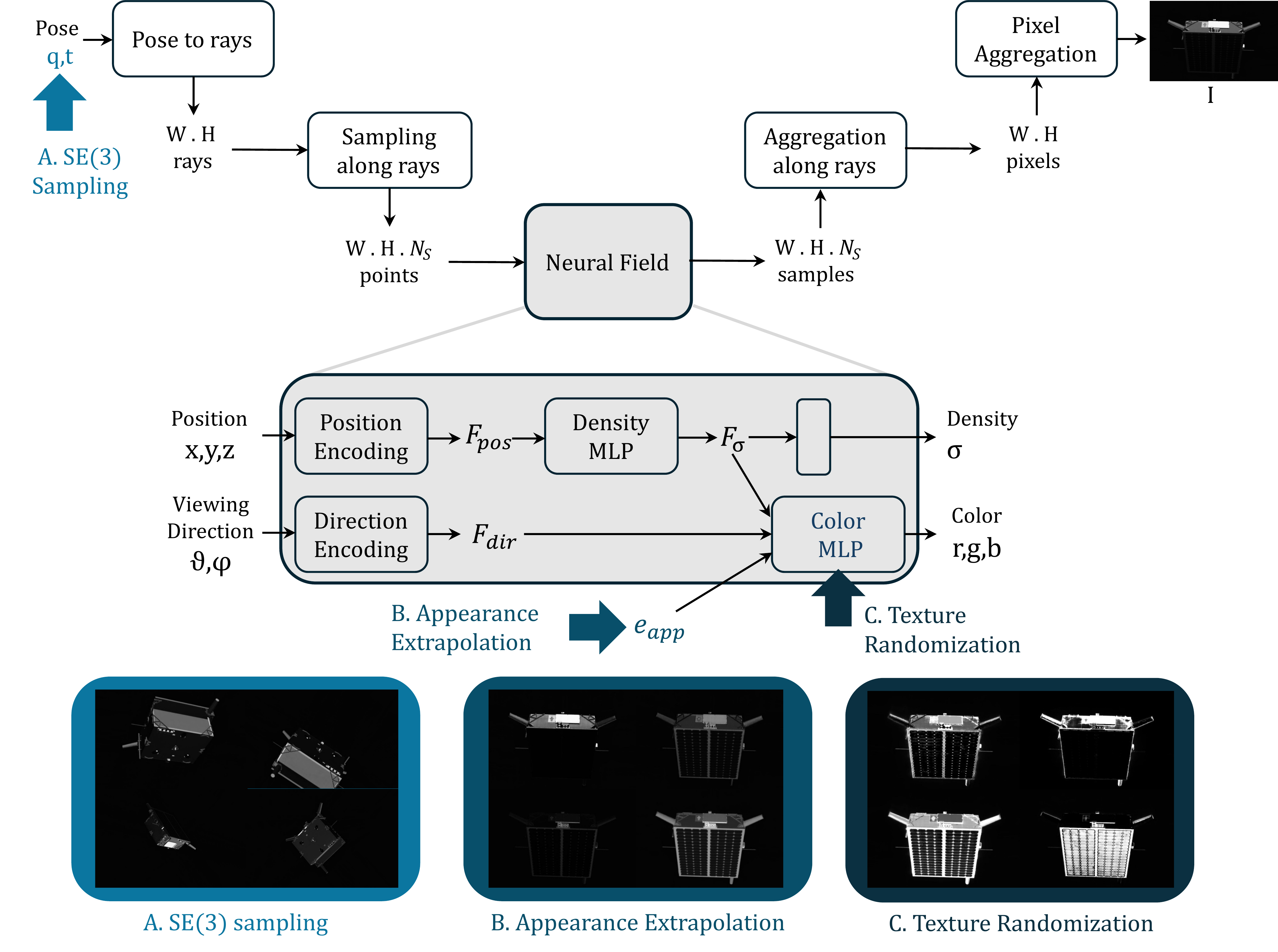}
    \caption{\label{fig_itw_nerf_overview} Overview of the synthesis of an image $I$ depicting an object under a camera pose ($q$,$t$) through an in-the-wild NeRF, using our augmentation method. As explained in \Cref{sec_bacground_wild_nerfs}, the scene is represented by a field that maps the coordinates of any point to its density and color. By aggregating those values along rays passing through each pixel of the camera, the image can be generated. Our augmentation method leverages the NeRF to enable the generation of images, (A) taken under novel viewpoints, (B) depicting rich illumination conditions, and (C) presenting diverse textures. This significantly improves the diversity of the generated set, which, in turn, enhances the generalization capabilities of a pose estimator trained on that set.}
\end{figure} 

    Neural Radiance Fields~\cite{mildenhall2021nerf} have recently appeared as an efficient tool for rendering novel views of a 3D scene. They represent a scene as a field, approximated through a neural network, which maps a 3D position and 2D viewing direction into a rgb color and a density. To render a novel view under a given camera pose, rays are shot through each pixel of that virtual camera. By querying the network, the color and density of points sampled along each ray can be estimated. Finally, the value of each pixel of the image can be recovered by aggregating the color and density of the points along the corresponding ray through differentiable ray-tracing techniques. Several improvements significantly enhanced the capabilities of NeRFs. Instant-NGP~\cite{muller2022instant} introduced a Multi-Resolution Hash encoding that reduced the typical training time from a few hours down to a few seconds. In-the-wild NeRF~\cite{martin2021nerf} added learnable appearance embeddings to the original NeRF to capture the varying appearance conditions that may exhibit two images of the same scene taken under the same viewpoints but at different times. Finally, $K$-Planes~\cite{fridovich2023k} combined the fast encoding of Instant-NGP with the appearance embeddings of in-the-wild NeRFs. In addition, they proposed a novel encoding of the position of the sampled points that makes the representation more explicit. 

    Several works explored the use of NeRFs for data augmentation. Neural-sim~\cite{ge2022neural} relied on a NeRF to generate training data for downstream tasks. It exploited a bi-level optimization to train both the downstream networks as well as the NeRF so that the synthesized images further improve the accuracy of the downstream networks on the validation set. As a result, the approach presents no interest for single domain generalization since the downstream networks tend to overfit the validation domain. Legrand~\etal\cite{legrand2024leveraging} exploited a Neural Radiance Field to extend a training set made of a few real images depicting an unknown object. However, they only exploited the NeRF to enable the training of pose estimator from a few real images and did not address the domain generalization problem. Unlike them, our works leverage a NeRF trained on synthetic images to augment the diversity of a training set in terms of viewpoints, illumination conditions and textures. This results in improved generalization abilities for a pose estimator trained on that set.

\section{Method}
\label{sec_method}

    This section describes our method for augmenting a training set to improve the generalization abilities of a 6D pose estimation network. 
    As highlighted by previous works~\cite{tobin2017domain,wang2021learning,legrand2024domain}, the diversity of the training set on which a network is trained is of considerable importance for its generalization abilities. Our method leverages this conclusion to improve the generalization capabilities of a network by training it on a diverse train set synthesized using a Neural Radiance Field. As illustrated in \Cref{fig_overview_intro}, $N_{synth}$ synthetic images depicting the object are picked from the synthetic set, $S_{synth}$, generated using the target CAD model, along with their pose labels. An in-the-wild NeRF~\cite{martin2021nerf}, is then trained on those images. The NeRF is exploited to synthesize an augmented set $S_{nerf}$ containing as many training images as required, rendered in diverse illumination and texture conditions. Finally, the pose estimator is trained on a train set $S_{train}$ which contains both the synthetic and augmented sets. 
    
    The following sections present the pipeline used to generate the augmented set. \Cref{sec_bacground_wild_nerfs} introduces the general architecture of a so-called in-the-wild NeRF~\cite{martin2021nerf}, while \Cref{sec_nerf_augm} describes how the rendering pipeline can be adapted to increase the diversity of the generated images and therefore improve the generalization capability of the downstream pose estimation networks.

\begin{figure}[b]
    \centering
    \includegraphics[width=0.99\linewidth]{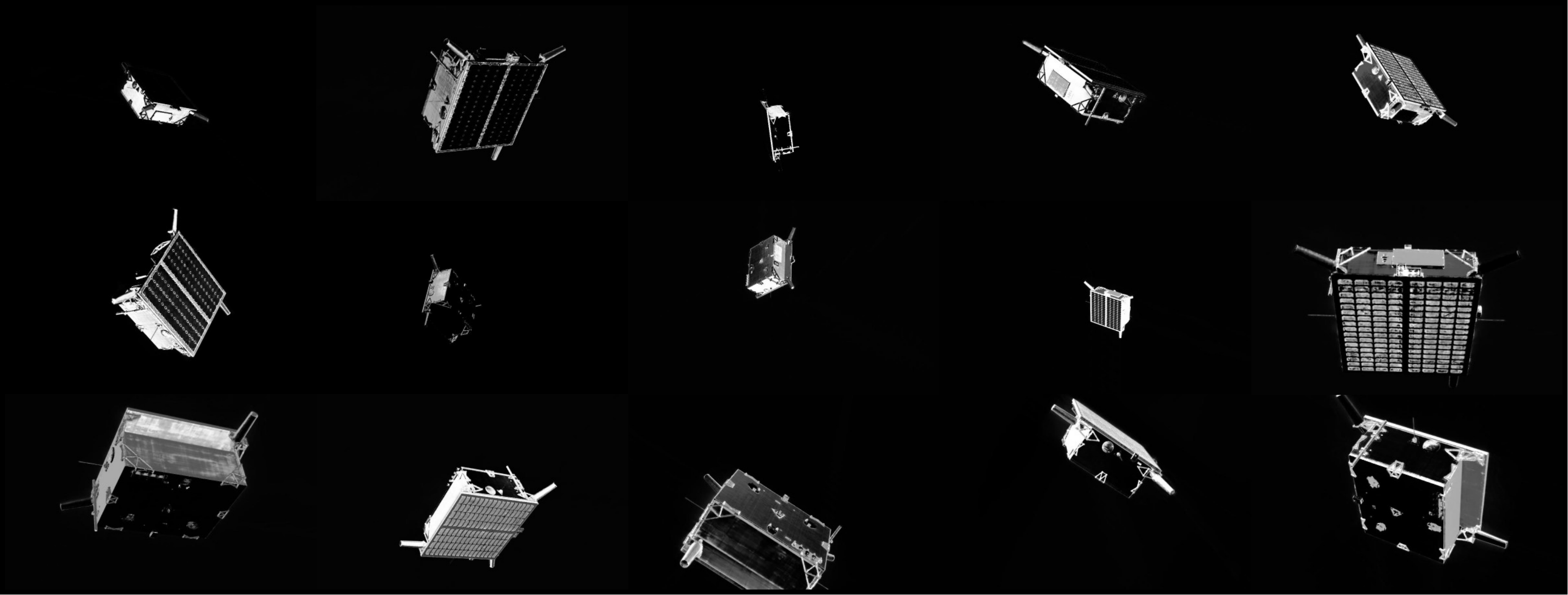}
    \caption{\label{fig_examples_combined} Examples of images generated by our NeRF-based augmentation method. While the synthetic images on which the NeRF was trained only depicts a texture-less target spacecraft exposed to smooth illumination conditions (see \Cref{fig_mismatches_domain}), our images exhibit a much larger diversity in terms of both illumination conditions and texture. In addition, our method enables the synthesis of images taken from novel viewpoints. For a comprehensive visual demonstration of this diversity, see the supplementary video.}
\end{figure}
    
\subsection{In-the-Wild Neural Radiance Fields}
\label{sec_bacground_wild_nerfs}

    \Cref{fig_itw_nerf_overview} depicts how to synthesize an image $I$, of resolution $W \times H$, observed by a camera of pose $(q,t)$, using an in-the-wild Neural Radiance Field~\cite{martin2021nerf}. Given the camera calibration matrix, the queried pose, \ie, position $t$ and rotation $q$, is decomposed into $H \cdot W$ rays passing by the center of each of pixel. Along each ray, $N_{samples}$ points are sampled. Each of those points is defined by a 3D position ($x$, $y$, $z$) and two viewing angles ($\theta$, $\phi$) in the scene referential. 
    
    For each point, the 3D position is mapped into position features $F_{pos}$ while the two viewing directions are mapped into direction features $F_{dir}$. For example, the position features can be predicted through a $K$-Planes encoding~\cite{fridovich2023k} while the direction features can be computed using spherical harmonics. However, the exact nature of those mappings does not affect our method, as long as position and direction features are predicted. The position features are then processed by a MLP that outputs density features $F_{\sigma}$ which are then used to predict the density $\sigma$ of the 3D point. Then, the density features $F_{\sigma}$ and direction features $F_{dir}$ are concatenated with an appearance embedding $e_{app}$. The resulting features are fed in a second MLP which predicts the $rgb$ color of the 3D point, viewed under the given direction, in the appearance conditions implicitly specified by the appearance embedding. 
    
    The image is then generated by aggregating for each pixel the points along the corresponding rays through ray-tracing techniques. While a usual NeRF is only able to represent a scene for which the training images were captured under the same appearance conditions, an in-the-wild NeRF can learn a scene for which those images were taken under the varying appearance conditions. This ability to capture the appearance observed in the wild directly comes the appearance embedding used by the color MLP.

\subsection{NeRF-based Augmentations}
\label{sec_nerf_augm}

    To increase the diversity of the initial synthetic training set, we leverage the concept of in-the-wild Neural Radiance Fields. This is done along three axes described below. 
    Those three augmentations can either be applied independently or jointly. 
    \Cref{fig_examples_combined} depicts some images that are generated using the three augmentations together. 
    
    \subsubsection{A. Coverage of the SE(3) space}
    As illustrated in \Cref{fig_itw_nerf_overview}, a first, trivial, augmentation enabled by the use of a NeRF is the synthesis of additional training images with novel pose labels. Indeed, since a NeRF can generate an image of a learned scene from arbitrary viewpoints, we can provide novel pose labels to the NeRF and improves the coverage of the SE(3) space through the resulting images. This augmentation does not contribute to the generalization abilities of the pose estimator but may address a data scarcity problem encountered in several use-cases~\cite{legrand2024leveraging}. This is further discussed in \Cref{sec_ablation_generation}.
    
    \subsubsection{B. Appearance Extrapolation}
    A second augmentation deals with the illumination conditions synthesized by the NeRF. As depicted in \Cref{fig_itw_nerf_overview}, an in-the-wild NeRF relies on an appearance embedding $e_{app}$ to control the object rendering illumination conditions. This embedding belongs to a D-dimensional space. For each of the $N$ images on which the NeRF is trained, a learnable appearance embedding is associated and trained along the NeRF. As a result, the appearance embedding can represent the illumination conditions that are specific to the corresponding image. 
    
\begin{figure}[t]
    \centering
    \includegraphics[width=1.0\linewidth]{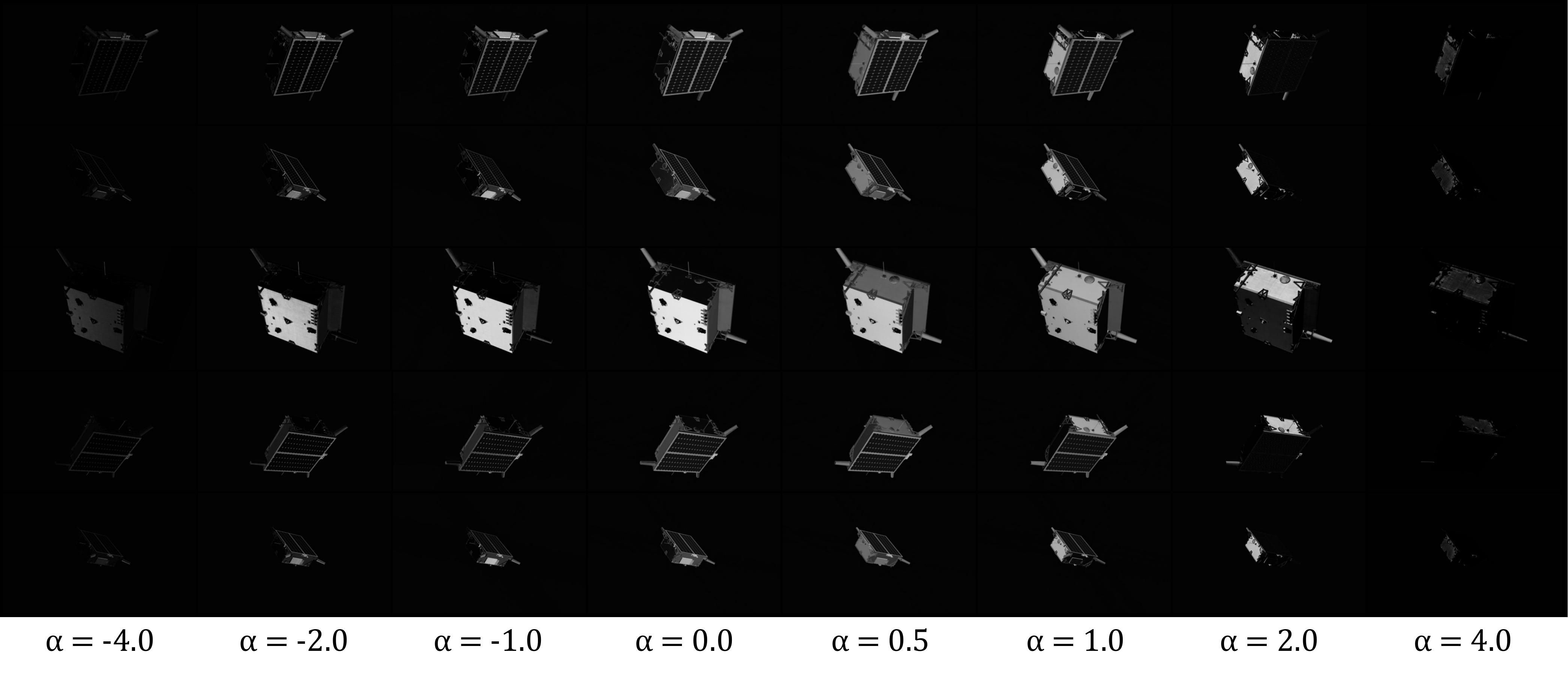}
    \caption{\label{fig_examples_app_extrapolation} Images generated through appearance interpolation/extrapolation. Each line depicts the same target, under the same pose, but using different weights $\alpha$ for the extrapolation of the appearance embedding. The illumination conditions are more diverse when the appearance is extrapolated ($\alpha \in [-4, 4]$) rather than only interpolated ($\alpha \in [0, 1]$). See the supplementary video for an example of a sequence generated through a progressive extrapolation between two appearance embeddings.}
\end{figure}

    Novel appearance embeddings can be sampled from the appearance space to increase the diversity of the illumination conditions. A straightforward augmentation would consist in randomly selecting one of the $N_{train}$ embeddings for each novel image. However, previous works~\cite{martin2021nerf} have shown that the appearance space is well structured. Therefore, we propose to interpolate between the learned embeddings, and even to extrapolate from them. Formally, for each novel image, we randomly pick two appearance embeddings out of the $N_{train}$ trained ones. Let $e_{i}$ and $e_{j}$ denote those two embeddings. A novel embedding $e_{app}$ is interpolated/extrapolated as
    \begin{equation}
        \label{eq_interpolation}
        e_{app} = e_{i} + \alpha (e_{j} - e_{i}),
    \end{equation}
    with $\alpha \in [0,1]$ for interpolation, and $\alpha \in [-4,0[ \cup ]1,4]$ for extrapolation.
    As depicted in \Cref{fig_examples_app_extrapolation}, using extrapolated appearance embeddings leads to images with more diverse illumination conditions compared to interpolated embeddings ($\alpha \in [0,1]$) or embeddings randomly picked from the trained ones ($\alpha$ equal to 0 or 1). 
    \Cref{sec_ablation_appearance} evaluates those different appearance strategies on the generalization abilities of a pose estimator.
       
\begin{figure}[t]
    \centering
    \includegraphics[width=1.0\linewidth]{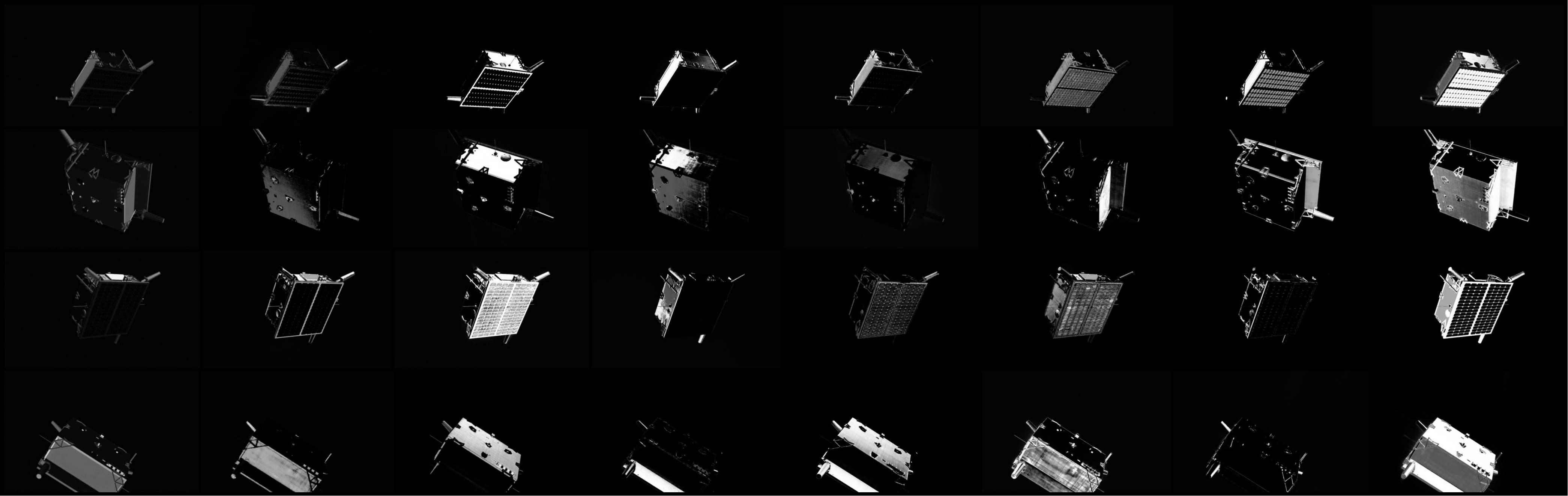}
    \caption{\label{fig_examples_texture_augmentation} Increasing the image texture diversity. Each line depicts the same target, under the same pose, but using randomized color MLP weights. Randomizing those weights does not affect the physical coherence of the target but modify its texture. For a comprehensive visual demonstration of the diversity enabled by this texture augmentation, see the supplementary video.}
\end{figure}

    \subsubsection{C. Texture Randomization}
    Finally, as a third augmentation, we introduce an original approach to enlarge the object texture diversity. This augmentation is motivated by the fact that previous works~\cite{geirhos2018imagenet,hendrycks2021many} have shown that Convolutional Neural Networks (CNNs) as the ones involved in the pose estimator, tend to overfit the texture. Given the mismatches between synthetic and real images (see \Cref{fig_mismatches_domain}), this overfitting significantly impairs the generalization abilities of a pose estimator~\cite{park2024robust,legrand2024domain}. Hence, we propose to increase the diversity of the train set by playing on the object texture.

    This texture diversification requires to distort the visual appearance of the target without affecting its geometry. As illustrated in \Cref{fig_itw_nerf_overview}, the geometry of the object only depends on the density of the points sampled in the neural field. Hence, as long as the density of those points is preserved, the 3D consistency of the target is ensured. Therefore, to increase texture diversity, we propose to add random noise on the weights of the color MLP, which is expected to randomize the predicted color. Empirically, we found that applying a Gaussian noise of zero-mean and standard deviation of 4.0 enables the generation of images with diverse textures, as depicted in \Cref{fig_examples_texture_augmentation}. \Cref{sec_ablation_texture} discusses the benefits of this texture augmentation on the generalization abilities of a pose estimator.

\section{Experiments}
\label{sec_experiments}

    This section evaluates the benefits brought by our NeRF-based image synthesis method on the generalization abilities of a spacecraft pose estimation network trained on a set augmented by our approach. \Cref{sec_expe_setup} describes our experimental setup. \Cref{sec_expe_main} evaluates the improvement of the pose estimator generalization abilities when using our image synthesis method to augment an original synthetic set. Sections \ref{sec_ablation_appearance} and \ref{sec_ablation_texture} explore the impact on the pose estimator generalization abilities of the appearance extrapolation and texture randomization strategies, respectively. Finally, \Cref{sec_ablation_generation} demonstrates that our nerf-based set can be used to train a robust pose estimator on its own.
    
\subsection{Experimental setup}
\label{sec_expe_setup}

    The experiments were conducted on the SPEED+~\cite{park2022speed+} dataset which was specifically designed to study the impact of the domain shift on spacecraft pose estimators. It contains both synthetic and Hardware-In-the-Loop (HIL) grayscale images depicting TANGO, the target spacecraft of the PRISMA mission~\cite{gill2007autonomous}, with the corresponding pose labels. \Cref{tab_datasets} presents an overview of the different sets belonging to SPEED+~\cite{park2022speed+}. The \textit{Synthetic} set contains 59,960 images synthesized using an OpenGL based rendering pipeline. On half of the set, real images depicting the Earth were added in the background. The distance between the spacecraft and the camera ranges from 2.2 to 10 meters. In our experiments, 48,000 images ($80\%$) of the \textit{Synthetic} set are used for training while the rest of the set is only used for validation. From the domain generalization perspective, the \textit{Synthetic} set corresponds to the source domain while the two HIL sets correspond to target domains. Those were captured in the TRON facility~\cite{park2021robotic} which reproduces the lighting conditions observed in Low Earth Orbit, using a mock-up of the target spacecraft and a space-grade camera. Each of the HIL sets mimics different illumination scenarios. The \textit{Sunlamp} set reproduces the lighting conditions encountered when the Sun directly lights the spacecraft, while the \textit{Lightbox} set mimics a more diffuse illumination, as the one encountered when the spacecraft is lit by the Earth albedo. The \textit{Lightbox} and \textit{Sunlamp} sets contain 6740 and 2791 images, respectively. In both HIL sets, the distance between the target and the camera ranges from 2.5 to 9.5 meters.
        
 \begin{table}[t]
    \caption{\label{tab_datasets} Overview of the SPEED+ dataset~\cite{park2022speed+}. It contains three sets. The \textit{Synthetic} set contain images generated by a rendering tool which does not properly render the illumination conditions, nor the spacecraft texture. The \textit{Lightbox} and \textit{Sunlamp} sets were captured in a testbed that emulates the lighting conditions encountered on orbit and uses a realistic mock-up of the target. The illumination conditions in \textit{Lightbox} correspond to a diffuse illumination, \eg, the Earth albedo, while, in \textit{Sunlamp}, they correspond to a direct illumination of the spacecraft by the Sun.}
    \centering
    \setlength\tabcolsep{2pt} 
    \renewcommand{\arraystretch}{1.1}
    \begin{tabular}{|c|c|c|c|}
        \hline
        SPEED + & \textit{Synthetic} & \textit{Lightbox} & \textit{Sunlamp}\\
        \hline
        Domain & Synthetic (Source) & \multicolumn{2}{c|}{Testbed (Target)}\\
        \hline
        Illumination & Synthetic & Diffuse & Direct \\
        \hline
        Spacecraft & Simplified model & \multicolumn{2}{c|}{Realistic mockup} \\
        \hline
        & \includegraphics[width=0.27\linewidth]{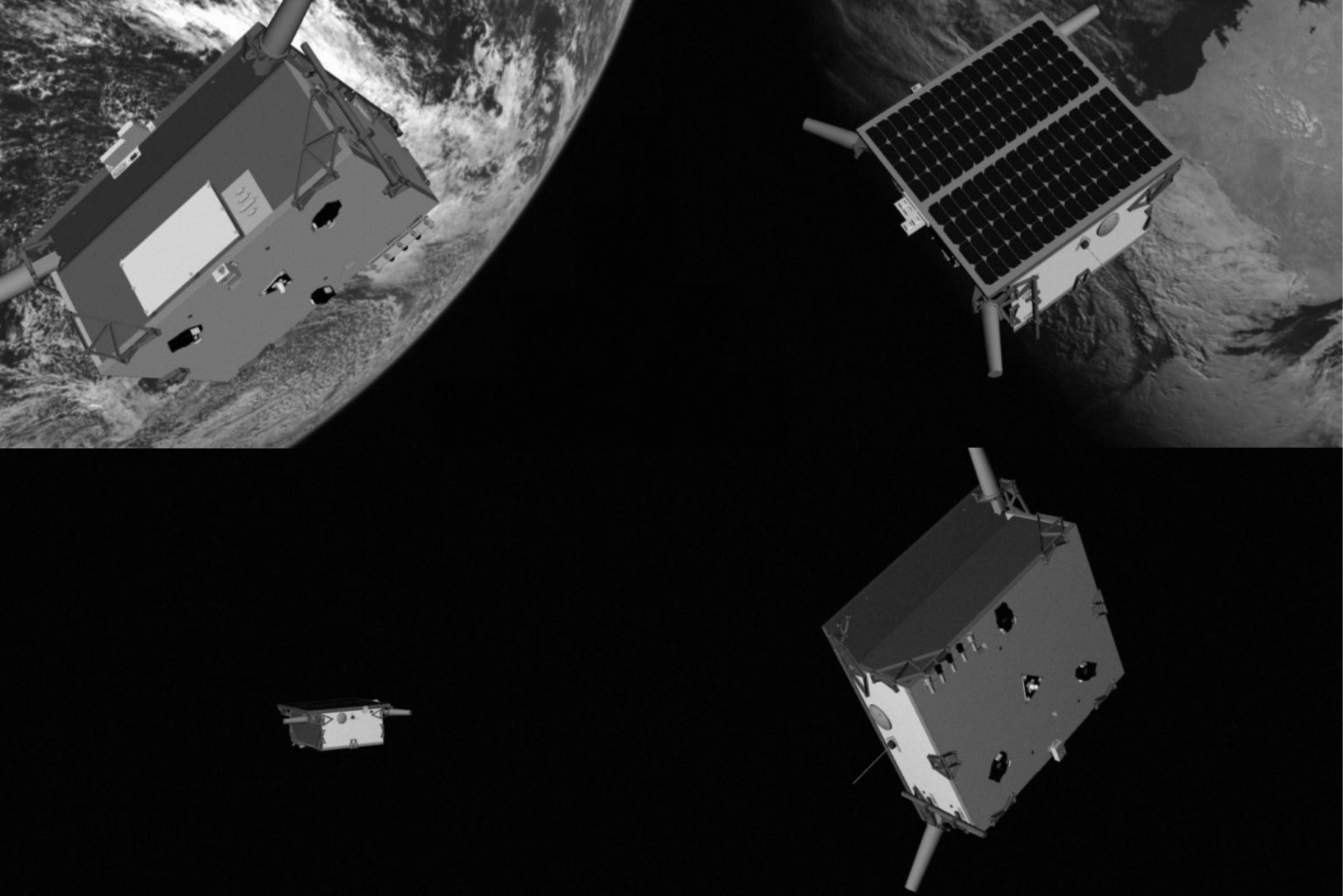} & \includegraphics[width=0.27\linewidth]{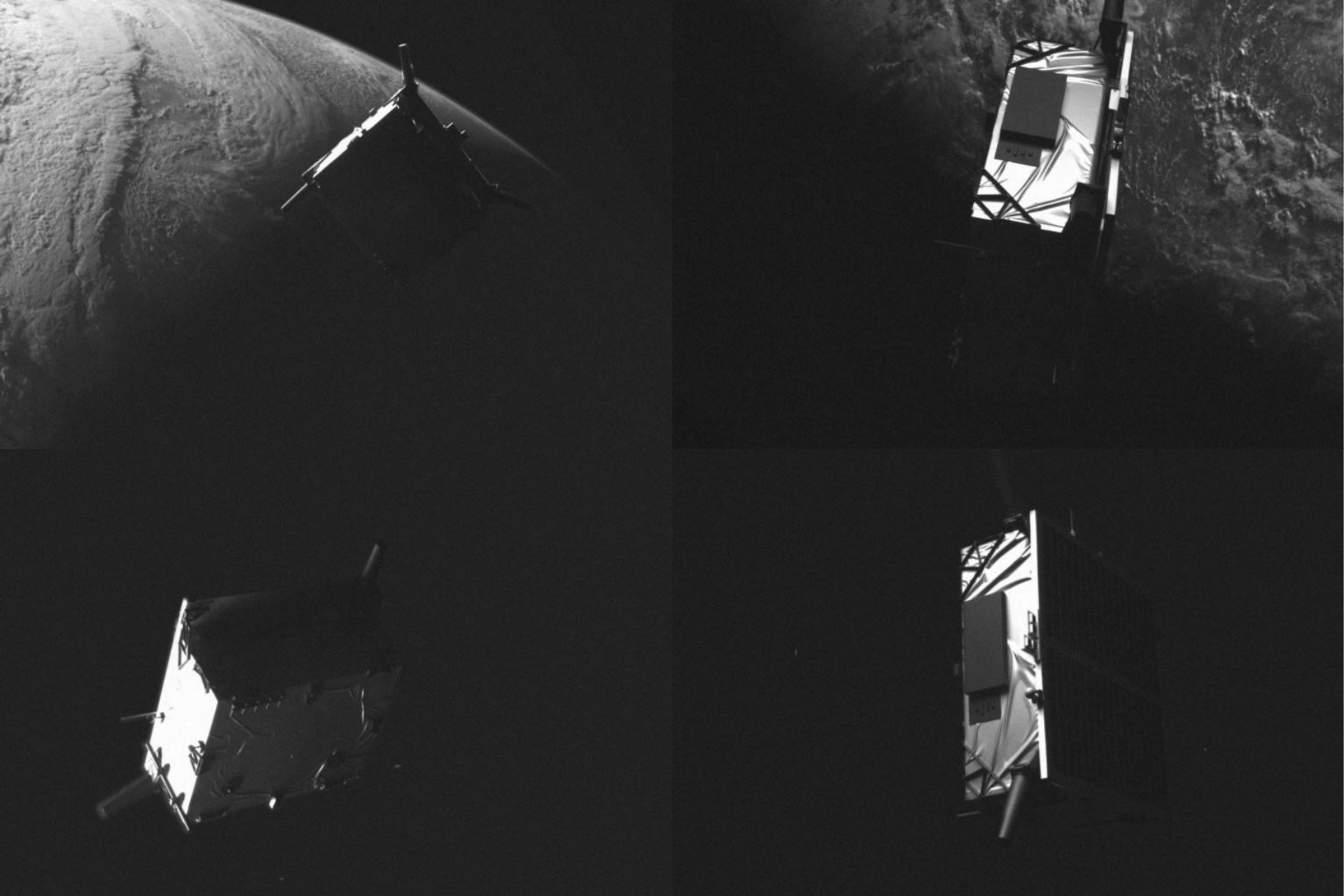} & \includegraphics[width=0.27\linewidth]{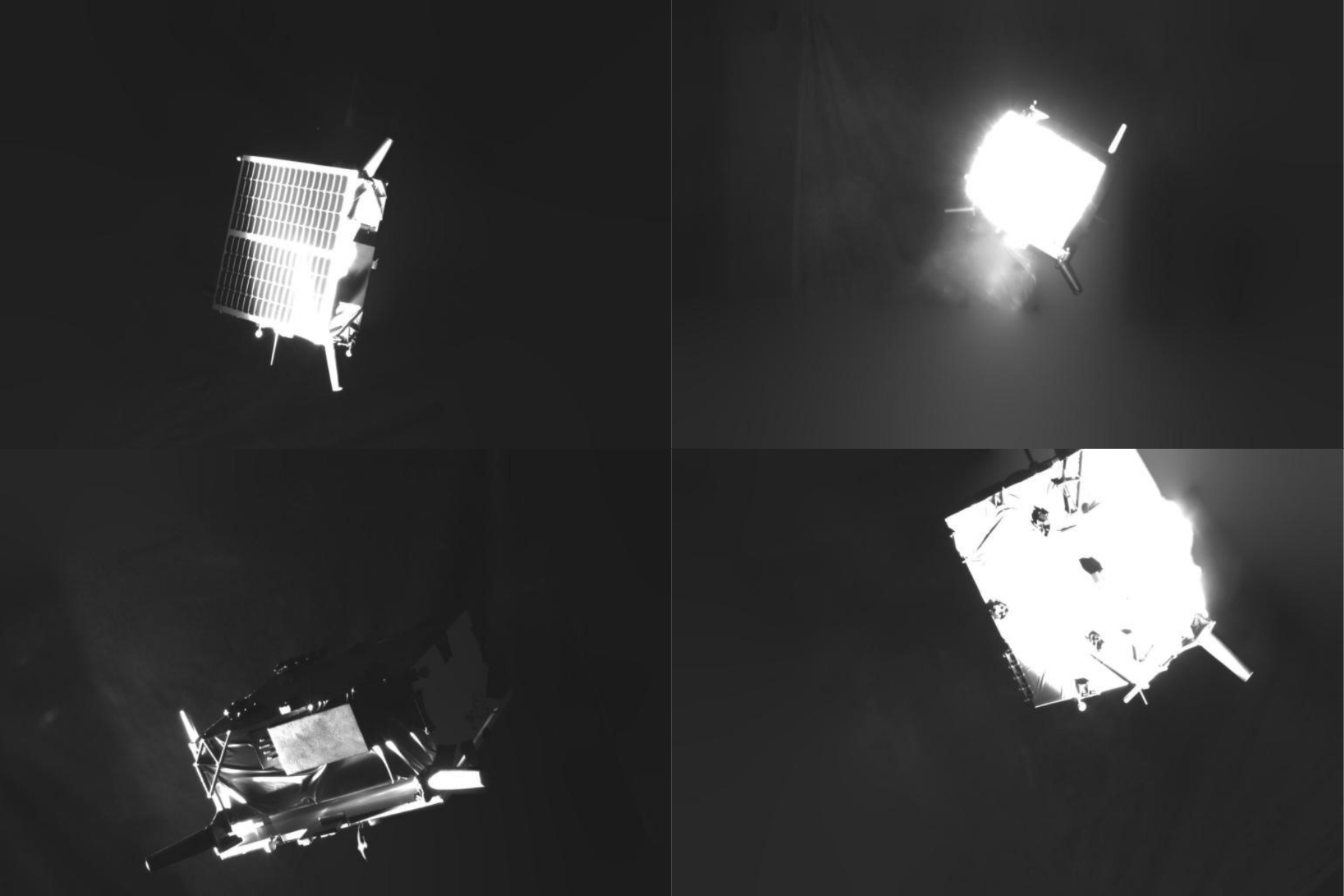} \\
        \hline
    \end{tabular}    
\end{table}

    Following common practices on this dataset~\cite{park2023satellite,perez2022spacecraft,wang2023bridging}, the pose estimator is evaluated using the SPEED+ score~\cite{park2022speed+}, $S^*_{pose}$. It sums up the angular error $E_{R}$ (in radians) with the normalized translation error $E_{TN}$. We also compute the translation error $E_{T}$ in centimeters. All the values in this section are averaged on the considered test set.

    As explained in \Cref{sec_method}, our method relies on a train set $S_{train}$, which combines an original synthetic set $S_{synth}$ with an augmented set $S_{nerf}$ synthesized through a Neural Radiance Field. We train this NeRF using 500 images of the \textit{Synthetic} set with a $90\%-10\%$ train-val split. 
    $N_{nerf}$ pose labels are sampled from the SE(3) space. For each sample, we generate two images. The first is synthesized using only the appearance extrapolation augmentation while the second is rendered using in addition the texture randomization technique described in \Cref{sec_nerf_augm}. All the images are converted in grayscale. Our NeRF implementation follows the $K$-Planes~\cite{fridovich2023k} one. Training the NeRF takes 40 minutes on a NVIDIA RTX3090. The rendering of a batch of 1000 images, of resolution 960 per 600 pixels, takes 20 minutes on a single NVIDIA RTX3090.

\begin{figure}[b]
    \centering
    \includegraphics[width=0.75\linewidth]{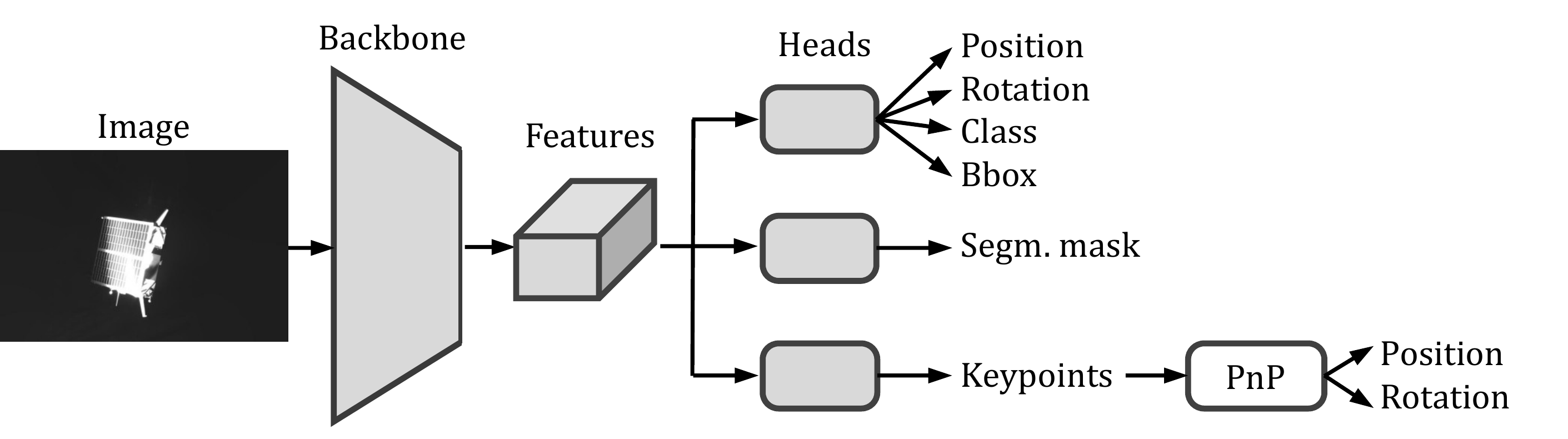}
    \caption{\label{fig_spnv2} Overview of the pose estimation network, SPNv2~\cite{park2024robust}. It extracts features from the image through a backbone made of an EfficientNet~\cite{tan2019efficientnet} encoder followed by BiFPN~\cite{tan2020efficientdet} layers. The features are used by multiple heads. The first heads form an EfficientPose~\cite{bukschat2020efficientpose} head that predicts the target position, orientation, class, and bounding box. A second head outputs a segmentation mask while the third one predicts $K$ heatmaps which highlight $K$ pre-defined keypoints. Their coordinates are processed by a PnP solver~\cite{lepetit2009ep} to predict the target position and orientation. At inference, only the position predicted by the EfficientPose head, and the keypoint-based rotation are considered.}
\end{figure}

    In the following experiments, the pose estimation network is SPNv2~\cite{park2024robust}. This pose estimation network was specifically designed for estimating the pose of a target spacecraft. As depicted in \Cref{fig_spnv2}, it relies on a backbone to extract features that are processed by multiple heads. This multi-task learning strategy, combined with image augmentation strategies, has been shown to significantly improve the generalization abilities of the network~\cite{park2024robust}. This network was chosen because it achieves state-of-the-art generalization abilities on the spacecraft pose estimation task. In this paper, we show that generalization can be improved even further through our NeRF-based augmentation approach.

    The pose estimator is trained on a set $S_{train}$ that combines the synthetic images of the \textit{Synthetic} set ($S_{synth}$) with the generated ones ($S_{nerf}$). Hence, it is trained for 20 epochs made of 48,000 + $N_{nerf}$ images, with a batch size of 16 images of resolution 768$\times$512. Unless otherwise mentioned, $N_{nerf}$ is set to 48,000. For each nerf-based training example, we randomly sample the image generated either using only the appearance extrapolation strategy or combining it with texture randomization. The training images undergo the same augmentation pipeline as in the SPNv2 paper~\cite{park2024robust}, including neural style augmentation~\cite{jackson2019style}. In addition, we add the Earth in the background of half of the images, as in~\cite{park2022speed+}.

\subsection{NeRF-based Augmentation Benefit}
\label{sec_expe_main}

    \Cref{tab_main_results} compares the accuracy achieved by a pose estimation network, \ie, SPNv2, trained either on the \textit{Synthetic} set, $S_{synth}$, of SPEED+, or on a set $S_{train}$, combining $S_{synth}$ with $S_{nerf}$, \ie, a set generated by our method using both appearance extrapolation and texture randomization. It highlights that our method significantly improves the generalization abilities of the pose estimation network. The average errors are decreased by $55\%$ and $45\%$ on the \textit{Lightbox} and \textit{Sunlamp} sets, respectively.

\begin{table*}[h]
\centering
\caption{\label{tab_main_results} Average performance metrics achieved on \textit{Lightbox} and \textit{Sunlamp} by the SPNv2 pose estimation network trained either on the \textit{Synthetic} set ($S_{synth}$) or on a train set ($S_{train}$) that combines $S_{synth}$ with our nerf-based generated set ($S_{nerf}$). Our augmentation technique significantly improves the generalization abilities of the pose estimator.}
\begin{tabular}{c|rrr|rrr}
    \toprule
    Strategy & \multicolumn{3}{c|}{\textit{Lightbox}} & \multicolumn{3}{c}{\textit{Sunlamp}} \\
     & $S^{*}_{Pose} [/]$ & $E_{R}[\degree$] & $E_{T}[cm]$ & $S^{*}_{Pose}[/]$ & $E_{R}[\degree]$ & $E_{T}[cm]$ \\ 
    \midrule
    Baseline & 0.191 & 8.88 & 23.1 & 0.305 & 14.90 & 28.6 \\
    Ours & \textbf{0.085} & \textbf{3.66} & \textbf{14.1} & \textbf{0.169} & \textbf{7.82} & \textbf{21.4} \\
    \bottomrule
\end{tabular}
\end{table*}

\subsection{Ablation Study: Appearance Extrapolation}
\label{sec_ablation_appearance}
    Our NeRF-based augmentation method proposes to extrapolate the appearance embedding $e_{app}$ (see \cref{fig_itw_nerf_overview}) to enrich the illumination conditions of the generated set. This section demonstrates the importance of this strategy on the generalization abilities of the pose estimation network. \Cref{tab_ablation_appearance} compares the performance achieved by a pose estimator trained on sets augmented with different illumination augmentation strategies. Interpolating between the appearance embeddings does not perform as well as randomly selecting them. This comes from an oversampling of the average appearance conditions which impairs the diversity of the generated set. Interestingly, extrapolating the appearance embeddings reduces the average errors by $21\%$ and $36\%$ on \textit{Lightbox} and \textit{Sunlamp}, respectively, compared to using randomly picked appearance embeddings. This demonstrates that extrapolating the appearance is crucial for the diversity required by the pose estimator train set.

\begin{table*}[h]
\centering
\caption{\label{tab_ablation_appearance} Average performance metrics achieved on \textit{Lightbox} and \textit{Sunlamp} by the SPNv2 pose estimation network trained on a train set augmented using as appearance augmentation strategy either, (a) randomly picked appearance embeddings, (b) appearance embeddings interpolated between two randomly picked embeddings, (c) extrapolated between those two embeddings with a weight $\alpha$ in \cref{eq_interpolation} picked between -1 and 1, or (d) extrapolated but using a weight $\alpha$ picked between -4 and 4. The appearance extrapolation strategy improves the domain generalization abilities of the pose estimator.}
\begin{tabular}{c|rrr|rrr}
    \toprule
    Appearance Augm. & \multicolumn{3}{c|}{\textit{Lightbox}} & \multicolumn{3}{c}{\textit{Sunlamp}} \\
    Strategy & $S^{*}_{Pose} [/]$ & $E_{R}[\degree$] & $E_{T}[cm]$ & $S^{*}_{Pose}[/]$ & $E_{R}[\degree]$ & $E_{T}[cm]$ \\ 
    \midrule
    Random & 0.124 & 5.76 & 15.1 & 0.280 & 13.71 & 26.4 \\
    Interpolation $\alpha \in [0,1]$ & 0151 & 7.03 & 17.7 & 0.311 & 15.52 & 26.0 \\
    Extrapolation $\alpha \in [-1,1]$ & 0.112 & 5.19 & \textbf{13.8} & 0.249 & 11.92 & 25.3 \\
    Extrapolation $\alpha \in [-4,4]$ & \textbf{0.098} & \textbf{4.38} & 14.3 & \textbf{0.180} & \textbf{8.55} & \textbf{19.8} \\
    \bottomrule
\end{tabular}
\end{table*}

\subsection{Ablation Study: Texture Augmentation}
\label{sec_ablation_texture}
    \Cref{tab_ablation_Texture} evaluates the generalization abilities improvement brought by the combination of the texture randomization augmentation (\Cref{sec_nerf_augm}.C) with the appearance extrapolation strategy (\Cref{sec_nerf_augm}.B). It shows that randomizing the target texture further reduces the average errors by $13\%$ and $6\%$ on the \textit{Lightbox} and \textit{Sunlamp} sets, respectively.

\begin{table*}[h]
\centering
\caption{\label{tab_ablation_Texture} Average performance metrics achieved on \textit{Lightbox} and \textit{Sunlamp} by the SPNv2 pose estimation network trained 
on a train set augmented either using only the appearance extrapolation strategy or combining it with texture randomization. Augmenting the training set through texture randomization further improves the generalization abilities of the pose estimator.}
\begin{tabular}{c|rrr|rrr}
    \toprule
    Texture & \multicolumn{3}{c|}{\textit{Lightbox}} & \multicolumn{3}{c}{\textit{Sunlamp}} \\
    Randomization & $S^{*}_{Pose} [/]$ & $E_{R}[\degree$] & $E_{T}[cm]$ & $S^{*}_{Pose}[/]$ & $E_{R}[\degree]$ & $E_{T}[cm]$ \\ 
    \midrule
    \ding{55} & 0.098 & 4.38 & 14.3 & 0.180 & 8.55 & \textbf{19.8} \\
    \ding{51} & \textbf{0.085} & \textbf{3.66} & \textbf{14.1} & \textbf{0.169} & \textbf{7.82} & 21.4 \\
    \bottomrule
\end{tabular}
\end{table*}

\subsection{Future Use-Cases Perspectives}
\label{sec_ablation_generation}

    In previous sections, we showed that our method improves the generalization abilities of a pose estimator through the augmentation of a synthetic set $S_{synth}$ generated using the target CAD model. However, in some use-cases, such as space debris de-orbiting~\cite{forshaw2016removedebris}, this CAD model is not available and the information on the target may be limited to a few images~\cite{legrand2024leveraging}. This perspective motivates the results presented in \Cref{tab_ablation_generation}, which compares the accuracy of the pose estimator trained either on the 48,000 synthetic images of SPEED+~\cite{park2022speed+} (baseline), or on 48,000 images generated by a NeRF trained on 500 synthetic images of SPEED+. Our method outperforms the baseline although it uses 100 times fewer synthetic images. This preliminary result shows that our NeRF-based synthesis method is able to exploit the information contained in a limited number of images to generate a training set that is sufficiently rich to train an accurate pose estimator. This clearly opens the door for considering our NeRF-based approach in use-cases where the pose estimator has to be trained from a few real images, without access to a known target CAD model.
    
    
\begin{table*}[h]
\centering
\caption{\label{tab_ablation_generation} Average performance metrics achieved on \textit{Lightbox} and \textit{Sunlamp} by the SPNv2 pose estimation network trained either directly on a set generated by our method ($S_{nerf}$), or on the whole synthetic set ($S_{synth}$). Our approach, which only relies on a fraction ($1\%$) of the synthetic set to train the NeRF, outperforms the baseline which exploits the whole synthetic set.}
\begin{tabular}{l|r|rrr|rrr}
    \toprule
    Strategy & $S_{synth}$ & \multicolumn{3}{c|}{\textit{Lightbox}} & \multicolumn{3}{c}{\textit{Sunlamp}} \\
     & used $[\%]$ & $S^{*}_{Pose} [/]$ & $E_{R}[\degree$] & $E_{T}[cm]$ & $S^{*}_{Pose}[/]$ & $E_{R}[\degree]$ & $E_{T}[cm]$ \\ 
    \midrule
    Baseline & 100 & 0.191 & 8.88 & \textbf{23.1} & 0.305 & 14.90 & \textbf{28.6} \\
    \textbf{Ours} & \textbf{1} & \textbf{0.185} & \textbf{8.24} & 28.1 & \textbf{0.238} & \textbf{11.06} & 30.0 \\
    \bottomrule
\end{tabular}
\end{table*}

\section{Conclusion}
\label{sec_conclusion}

    This work introduced a novel data augmentation method for improving the generalization abilities of a 6D pose estimation network. Our method relies on a Neural Radiance Field, trained from synthetic images, which augments the train set of a pose estimator. This augmentation enriches the diversity of the train set in terms of viewpoints distribution, illumination conditions and textures. We validated the approach on a challenging use-case, the estimation of the relative pose of an uncooperative spacecraft. Due to the illumination and texture mismatches between the synthetic images on which pose estimation networks are trained and the real ones, those networks encounter a severe domain shift. We demonstrated on the SPEED+~\cite{park2022speed+} that our augmentation method successfully improves the generalization abilities of a spacecraft pose estimation network. In particular, our proposed appearance extrapolation and texture randomization strategies significantly increase the gain in test accuracy. Finally, we showed that the set of images generated by the NeRF is sufficient to train a robust pose estimation network on its own, which opens positive perspectives regarding the use of our approach in scenarios where the NeRF would be trained from a small set of real images, instead of trained on CAD-based synthetic images.

\footnotesize
\section*{Acknowledgements}
The research was funded by Aerospacelab and the Walloon Region through the Win4Doc program. Christophe De Vleeschouwer is a Research Director of the Fonds de la Recherche Scientifique - FNRS. Computational resources have been provided by the Consortium des Équipements de Calcul Intensif (CÉCI), funded by the Fonds de la Recherche Scientifique de Belgique (F.R.S.-FNRS) under Grant No. 2.5020.11 and by the Walloon Region.

\bibliographystyle{unsrt}  
\bibliography{references}

\end{document}